\documentclass{article}

\usepackage[dblblindworkshop, final]{neurips_2025}

\workshoptitle{Generative and Protective AI for Content Creation}

\usepackage[utf8]{inputenc}
\usepackage[T1]{fontenc}
\usepackage{hyperref}
\usepackage{url}
\usepackage{booktabs}
\usepackage{amsfonts}
\usepackage{nicefrac}
\usepackage{microtype}
\usepackage{xcolor}
\usepackage{graphicx}
\usepackage{wrapfig}
\usepackage{xparse}
\usepackage{xspace}
\usepackage{algorithm}%
\usepackage{fixltx2e}
\usepackage{tikz}
\usepackage{float}
\usepackage{multirow}
\usepackage{multicol}
\usepackage{colortbl}
\usepackage{array}
\newcommand{\PreserveBackslash}[1]{\let\temp=\\#1\let\\=\temp}
\newcolumntype{C}[1]{>{\PreserveBackslash\centering}p{#1}}
\newcolumntype{R}[1]{>{\PreserveBackslash\raggedleft}p{#1}}
\newcolumntype{L}[1]{>{\PreserveBackslash\raggedright}p{#1}}

\usepackage{microtype}
\usepackage{graphicx}
\usepackage{mathtools}
\usepackage{float}
\usepackage{subcaption}
\usepackage{booktabs} %
\usepackage[shortlabels]{enumitem}

\usepackage{amssymb}%
\usepackage{pifont}%

\usepackage{bbold}

\usepackage{hyperref,amssymb,enumitem}

\setlist[itemize]{leftmargin=*}
\setlist[enumerate]{leftmargin=*}

\renewcommand{\P}[1]{\operatorname{\mathbb{P}}[#1]}
\newcommand*{\rej}{{\ooalign{\lower.3ex\hbox{$\sqcup$}\cr\raise.4ex\hbox{$\sqcap$}}}}

\usepackage{stackengine}

\usepackage{xcolor}

\newcommand{\ie}{\textit{i.e.,}\@\xspace}

\DeclareRobustCommand\encircle[1]{\tikz[baseline=(char.base)]{\node[shape=circle,fill,inner sep=1pt] (char) {\textcolor{white}{#1}}}}

\graphicspath{{images/}}

\usepackage{booktabs,arydshln}
\makeatletter
\def\adl@drawiv#1#2#3{%
        \hskip.5\tabcolsep
        \xleaders#3{#2.5\@tempdimb #1{1}#2.5\@tempdimb}%
                #2\z@ plus1fil minus1fil\relax
        \hskip.5\tabcolsep}
\newcommand{\cdashlinelr}[1]{%
  \noalign{\vskip\aboverulesep
           \global\let\@dashdrawstore\adl@draw
           \global\let\adl@draw\adl@drawiv}
  \cdashline{#1}
  \noalign{\global\let\adl@draw\@dashdrawstore
           \vskip\belowrulesep}}
\makeatother

\usepackage{cleveref}
\crefalias{prop}{proposition} %

\newcommand{\nlp}[1]{}

\newcolumntype{x}[1]{>{\centering\arraybackslash\hspace{0pt}}p{#1}}

\usepackage{amsmath}
\usepackage{xspace}

\newcommand{\mink}{\textsc{Min-k\% Prob}\xspace}

\renewcommand{\P}{\textbf{P}\xspace}
\newcommand{\U}{\textbf{U}\xspace}

\newif\ifdraft

\ifdraft
\usepackage[textsize=tiny]{todonotes}
\newcommand{\janek}[1]{\textcolor{violet}{[JD: #1]}}
\newcommand{\mytodo}[1]{\textcolor{red}{[todo: #1]}}
\newcommand{\mycomment}[1]{\textcolor{red}{[comment: #1]}}
\newcommand{\antoni}[1]{\textcolor{magenta}{AK: #1}}
\newcommand{\adam}[1]{\textcolor{cyan}{[AD: #1]}}
\newcommand{\franzi}[1]{\textcolor{brown}{FB: #1}}

\else
\usepackage[disable]{todonotes}
\newcommand{\janek}[1]{}
\newcommand{\mytodo}[1]{}
\newcommand{\mycomment}[1]{}
\newcommand{\antoni}[1]{}
\newcommand{\adam}[1]{}
\newcommand{\franzi}[1]{}
\fi

\usepackage{amsmath,amsfonts,bm}

\def\eqref#1{equation~\ref{#1}}

\def\1{\bm{1}}

\DeclareMathAlphabet{\mathsfit}{\encodingdefault}{\sfdefault}{m}{sl}
\SetMathAlphabet{\mathsfit}{bold}{\encodingdefault}{\sfdefault}{bx}{n}

\title{Membership and Dataset Inference Attacks on Large Audio Generative Models}

\author{%
  \textbf{Jakub Proboszcz}\thanks{Equal contribution.} \\
  Warsaw University of Technology \\
  \And
  \textbf{Paweł Kochański}\footnotemark[1] \\
  Warsaw University of Technology \\
  \And
  \textbf{Karol Korszun}\footnotemark[1] \\
  Warsaw University of Technology \\
  \And
  \textbf{Donato Crisostomi} \\
  Sapienza University of Rome \\
  \And
  \textbf{Giorgio Strano} \\
  Sapienza University of Rome \\
  \And
  \textbf{Emanuele Rodolà} \\
  Sapienza University of Rome \\
  \And
  \textbf{Kamil Deja} \\
  Warsaw University of Technology \\
  IDEAS Research Institute \\
  \And
  \textbf{Jan Dubiński} \\
  Warsaw University of Technology \\
  NASK National Research Institute \\
  \texttt{jan.dubinski.dokt@pw.edu.pl} \\
}

\begin{document}

\maketitle

\begin{abstract}
Generative audio models, based on diffusion and autoregressive architectures, have advanced rapidly in both quality and expressiveness. This progress, however, raises pressing copyright concerns, as such models are often trained on vast corpora of artistic and commercial works. A central question is whether one can reliably verify if an artist’s material was included in training, thereby providing a means for copyright holders to protect their content. In this work, we investigate the feasibility of such verification through \textit{membership inference attacks} (MIA) on open-source generative audio models, which attempt to determine whether a specific audio sample was part of the training set. Our empirical results show that membership inference alone is of limited effectiveness at scale, as the per-sample membership signal is weak for models trained on large and diverse datasets. However, artists and media owners typically hold collections of works rather than isolated samples. Building on prior work in text and vision domains, in this work we focus on \textit{dataset inference} (DI), which aggregates diverse membership evidence across multiple samples. We find that DI is successful in the audio domain, offering a more practical mechanism for assessing whether an artist’s works contributed to model training. Our results suggest DI as a promising direction for copyright protection and dataset accountability in the era of large audio generative models.
\end{abstract}

\section{Introduction}

Generative audio models have undergone rapid advances in recent years, driven largely by diffusion (DMs)~\cite{tango,tango2, audioldm, audioldm2} and autoregressive architectures (ARMs)~\cite{figaro,audiogen}. These models are capable of producing highly realistic soundscapes, speech, and music. While this progress opens exciting opportunities in areas such as creative expression, accessibility, and interactive media, it also raises urgent concerns about privacy, copyright, and data governance. In particular, the vast datasets required to train such systems can contain artistic or commercial audio without transparent disclosure, leaving creators uncertain about whether their works contributed to a model’s capabilities~\cite{dornis2025generative}.
A central question in this context is whether one can reliably determine if a specific artist’s recordings were included in training of a generative model. Addressing this question is critical both for protecting intellectual property and for enabling accountability in machine learning practice. Similar challenges have been investigated in computer vision~\cite{kong2023efficient, dubinski2024towards, carlini2023extracting} and natural language processing~\cite{maini2024llmdatasetinferencedid, mattern2023membershipLLM, shi2024detecting}, where MIAs~\cite{shokri2017membershipinference} attempt to determine if a given sample was used in training, and DI~\cite{maini2021dataset, maini2024llmdatasetinferencedid, dubinski2025cdi} extends this idea to entire collections. However, the effectiveness of these techniques for large generative models in the audio domain remains unclear.

In this paper, we conduct a study of membership and dataset inference attacks on large audio generative models. We begin by evaluating the effectiveness of existing MIA strategies when applied to open-source ARMs and DMs. Our findings reveal that single-sample membership inference is weak in this setting, offering limited evidence of training set inclusion. Motivated by the observation that artists and rights-holders typically possess collections of works rather than isolated samples, we shift focus to DI. By aggregating diverse membership signals across multiple samples, DI achieves substantially higher effectiveness, enabling more reliable detection of training set participation. Our contributions are threefold:
\begin{itemize}[noitemsep,nolistsep]
\item We benchmark existing MIAs on large audio ARMs and DMs, highlighting their limitations.
\item We extend the existing DI methodology to audio generative models, assessing its effectiveness in the audio domain.
\item We provide an extensive empirical evaluation across several state-of-the-art audio models, demonstrating that DI can succeed where single-sample attacks fail, and thus suggest it as a promising mechanism for copyright protection and dataset accountability.
\end{itemize}

Our work aims to initiate a discussion on existing methods for protecting copyrighted audio samples in large-scale generative models, while also laying the groundwork for auditing methods that empower creators to assert control over their intellectual property.

\section{Background}

\subsection{Identifying Training Data}

\textbf{Membership Inference} (MIA).
MIAs~\citep{shokri2017membershipinference} aim to decide whether a given sample was part of a model’s training set. They exploit overfitting: training samples typically yield lower losses than unseen ones. Formally, the attacker constructs an attack function $A_{f_\theta}: \mathcal{X} \rightarrow {0,1}$ that predicts membership. A standard approach is the threshold attack~\citep{yeom2018lossmia}, which classifies $x$ as a member if the chosen metric is below a threshold: $A_{f_\theta}(x) = \mathbb{1}!\left[ \mathcal{M}(f_\theta, x) < \gamma \right]$, where $\mathcal{M}$ is the metric and $\gamma$ the decision threshold.

\textbf{Dataset Inference} (DI).
DI~\citep{maini2021dataset} asks whether an entire dataset was used during training. Unlike MIAs, which evaluate individual samples, DI aggregates membership signals (often based on MIAs) across multiple points into a dataset-level statistic, thereby amplifying weak per-sample evidence. This makes DI effective for large models and datasets where single-sample inference is unreliable. Initially proposed for supervised models, DI extracts per-sample features, aggregates them into a dataset score, and applies a statistical test~\citep{sslextractions2022icml,datasetinference2022neurips}. Recent work has extended DI to generative models, including large language models (LLMs)~\citep{maini2024llmdatasetinferencedid,zhao2025posthocDI}, DMs~\citep{dubinski2025cdi}, and autoregressive image models~\citep{kowalczukprivacy}. Formally, DI compares scores from a suspected member set and a non-member set via Welch’s $t$-test at $\alpha=0.01$ with the null hypothesis $H_0:\text{mean(member scores)} \leq \text{mean(non-member scores)}$. Rejecting $H_0$ implies the dataset was part of training. Correctness requires both sets be i.i.d.; otherwise, distributional mismatch can bias the test. The strength of DI depends on the number of available samples. To quantify leakage, we define $P$ as the minimum number of samples needed to reject $H_0$. Smaller $P$ indicates stronger leakage.

\begin{figure}[h]
    \centering
    \includegraphics[width=0.8\linewidth]{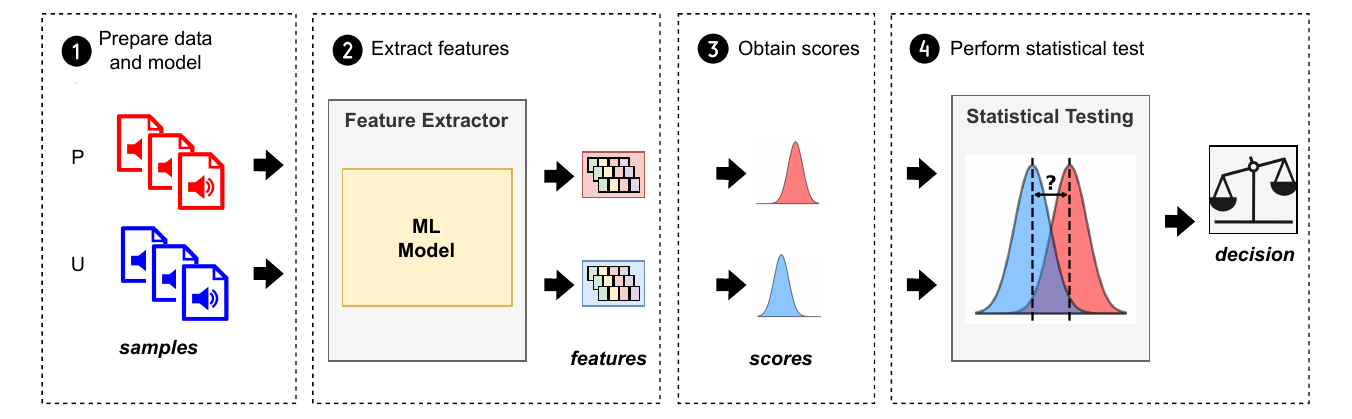}
\caption{\small \textbf{Dataset Inference Procedural Steps.} The process consists of four main steps: 
\encircle{1}~\textit{Data Preparation:} Prepare the data to verify whether the (suspected) member samples \P were used to train the model. The (confirmed) nonmember samples \U, from the same distribution as \P, serve as the validation set.
\encircle{2}~\textit{Feature Extraction:} Run each individual MIA on all inputs from $\{\P,\U\}$ to extract membership features for all data samples. We use our MIAs tailored to ARMs and DMs for the respective model types. 
\encircle{3}~\textit{Score Computation:} We map extracted feature vectors into scalar membership scores for each sample. We use a scoring model for DMs, following \cite{dubinski2025cdi}, and feature summation \cite{kowalczukprivacy} for ARMs (see Appendix for more details.
\encircle{4}~\textit{Statistical Testing:} Apply a statistical t-test to verify whether the scores obtained for the public suspect data points \P are statistically significantly higher than those for \U. If so, \P is marked as being part of the model's training set. Otherwise, the test is inconclusive and the model's training set is considered independent of \P.}
    \label{fig:di_schema}
\end{figure}

\subsection{Audio Generative Models}

\begin{table}[h!]
\centering
\scriptsize
\caption{\textbf{Comparison of audio generative models on which we experiment in our work.}}
\begin{tabular}{lcccccc}
\toprule
Model & Training steps & Dataset & Size & Params & Type & Input \\
\midrule
AudioGen   & 200k & AudioCaps + other & 4k h & 285M & AR & Audio tokens \\
FIGARO     & 100k & LakhMIDI & 176k files & 88M & AR & MIDI \\
AudioLDM2  & 800k & AudioCaps + other & 29.5k h & 1.5B & DM & Mel-spec \\
TANGO      & 230k & AudioCaps & 46k clips & 866M & DM & Mel-spec \\
\bottomrule
\end{tabular}
\label{tab:models}
\end{table}

We experiment on 4 models described in \Cref{tab:models} representing both AR and DM famielies. Currently, DMs dominate high-quality audio generation.  
\textbf{AudioLDM2}~\cite{audioldm2} unifies text-to-audio, text-to-music, and text-to-speech within one framework. It leverages a "Language of Audio" representation, mapped from AudioMAE~\cite{audiomae_huang2022amae} features through GPT-2~\cite{radford2019language}, to condition a UNet~\cite{ronneberger2015unet} diffusion model over mel-spectrogram latents. With 29.5k hours of diverse training data, it establishes a common semantic space that supports multiple generative tasks.  
\textbf{TANGO}~\citep{tango} follows a simpler design. A frozen FLAN-T5~\cite{chung2022scaling} encoder provides text embeddings that guide a latent diffusion model trained on the AudioCaps~\cite{audiocaps} dataset. Augmentation is based on pressure-level mixing of sounds, ensuring balanced exposure. Despite training on only 46k clips, TANGO achieves competitive generation quality by relying on strong instruction-tuned text features. 

Competing with DMs across different modalities, autoregressive approaches have also been applied to audio generation.  
\textbf{AudioGen}~\cite{audiogen} treats audio as a sequence of discrete codec tokens and learns a text-conditioned Transformer decoder to generate them. The model is trained on roughly 4k hours from ten heterogeneous audio–text corpora, and augmentations based on mixing sound sources are used to expose it to overlapping events.  \textbf{FIGARO}~\cite{figaro} addresses symbolic music with controllable generation. It introduces description-to-sequence learning, combining interpretable features such as chords, instrumentation, and rhythm with learned latent codes as conditioning. Training on the 176k-file LakhMIDI dataset~\cite{raffel2016learning} allows the model to reconstruct bar-level sequences and to provide global and fine-grained control.  

\section{Method}

\textbf{Membership Inference.}  
We begin by evaluating MIAs on large audio generative models. For DMs, we apply the attack suite explored in \cite{dubinski2025cdi}, which exploits denoising dynamics to distinguish training samples from non-members. For ARMs, we use the approach from Kowalczuk et al.~\cite{kowalczukprivacy}, which leverages token-level log-likelihoods and related statistics. We give more details on individual MIAs used in our work in the Appendix. In both cases, we use the train split of each model’s dataset as the source of \textit{members} and the held-out test split as the source of \textit{non-members}. This ensures that attacks are evaluated under a realistic and controlled setting where the attacker’s candidate pool contains both genuine training data and independent test samples.  

\textbf{Dataset Inference.}  
To facilitate the task, we extend our studies to DI. We follow the methodology introduced by \cite{dubinski2025cdi} for DMs and \cite{kowalczukprivacy} for ARMs. Each candidate dataset consists of a collection of suspected member samples $\mathcal{P}$ and an equal number of non-member samples $\mathcal{U}$ drawn from the test split. We extract membership features for each sample using multiple MIAs, aggregate them into scalar scores, and then apply Welch’s $t$-test to compare $\mathcal{P}$ and $\mathcal{U}$. The null hypothesis states that the mean score of $\mathcal{P}$ is no greater than that of $\mathcal{U}$, and we reject it at $\alpha=0.01$ if sufficient evidence is found. Following standard practice, we report the minimum number of samples $P$ required to reject the null hypothesis, with smaller $P$ indicating stronger information leakage.  Our approach is demonstrated on \Cref{fig:di_schema}.

\section{Results}
In our experiments on MIAs, we report the Area Under the Curve (AUC) and the True Positive Rate at a False Positive Rate of 1\% (TPR@FPR = 1\%). For AUC, a value of 0.50 corresponds to random guessing, while for TPR@FPR = 1\%, the baseline for random guessing is 0.01. For DI, we report the minimal number of samples in $\mathcal{P}$ required to successfully reject the null hypothesis $H_0$, \ie, to flag the audio samples in $\mathcal{P}$ as having been used in training a given model.

\setlength{\tabcolsep}{3pt} 

\begin{wraptable}{r}{0.6\linewidth}
\scriptsize
\centering
\caption{\textbf{MIA results for Autoregressive Models.} We report AUC and TPR@FPR=1\%.}
\label{tab:arm_mia_metrics}
\begin{tabular}{lcccc}
\toprule
& \multicolumn{2}{c}{\textbf{AUC}} & \multicolumn{2}{c}{\textbf{TPR@1\%}} \\
\cmidrule(lr){2-3}\cmidrule(lr){4-5}
\textbf{Attack} & AudioGen & FIGARO & AudioGen & FIGARO \\
\midrule
Loss~\cite{yeom2018lossmia}            & 52.85$\pm$0.00 & 50.28$\pm$0.61 & 0.68$\pm$0.00 & 1.18$\pm$0.11 \\
Zlib~\cite{carlini2021extractLLM}      & 50.46$\pm$0.00 & 49.37$\pm$0.61 & 0.74$\pm$0.00 & 1.11$\pm$0.22 \\
Hinge~\cite{bertran2024scalable}       & 54.42$\pm$0.00 & 50.05$\pm$0.60 & 1.42$\pm$0.00 & 1.12$\pm$0.21 \\
Min-K\%~\cite{shi2024detecting}        & 55.34$\pm$0.00 & 50.28$\pm$0.58 & 1.13$\pm$0.00 & 1.13$\pm$0.20 \\
Min-K\%$^{++}$~\cite{zhang2024min}     & 50.86$\pm$0.00 & 49.65$\pm$0.56 & 0.97$\pm$0.00 & 1.01$\pm$0.19 \\
CAMIA~\cite{chang2024context}          & 51.86$\pm$0.00 & 51.68$\pm$0.53 & 1.35$\pm$0.00 & 1.07$\pm$0.21 \\
\bottomrule
\end{tabular}
\vspace{-0.1cm}
\end{wraptable}

\Cref{tab:arm_mia_metrics} presents the AUC values and TPR@1\% for MIAs on ARMs. For both AudioGen and FIGARO, the results remain close to 50\% AUC, indicating chance-level performance. This suggests that single-sample membership inference is ineffective against SOTA audio ARMs, trained on larger datasets, as they do not leak strong per-sample signals. Similar observation can be seen with TPR at a fixed FPR of 1\%, where the values are consistently low, rarely exceeding 1\%, which confirms that MIAs provide limited evidence for distinguishing members from non-members in SOTA audio ARMs.

\begin{wraptable}{r}{0.6\linewidth}
\scriptsize
\vspace{-0.45cm}
\centering
\caption{\textbf{MIA results for Diffusion Models.} We report AUC and TPR@FPR=1\%.}
\label{tab:dm_mia_metrics}
\begin{tabular}{lcccc}
\toprule
& \multicolumn{2}{c}{\textbf{AUC}} & \multicolumn{2}{c}{\textbf{TPR@1\%}} \\
\cmidrule(lr){2-3}\cmidrule(lr){4-5}
\textbf{Attack} & AudioLDM2 & Tango & AudioLDM2 & Tango \\
\midrule
Loss~\cite{carlini2023extracting}              & 52.29$\pm$0.54 & 70.52$\pm$0.87 & 0.00$\pm$0.00 & 16.03$\pm$2.21 \\
Gradient Masking~\cite{dubinski2025cdi}  & 49.15$\pm$0.56 & 51.06$\pm$0.99 & 0.12$\pm$0.29 & 3.18$\pm$0.78 \\
Multiple Loss~\cite{dubinski2025cdi}     & 54.83$\pm$0.53 & 69.47$\pm$0.83 & 3.49$\pm$0.26 & 16.68$\pm$2.17 \\
NoiseOpt~\cite{dubinski2025cdi}          & 52.44$\pm$0.55 & 51.69$\pm$0.98 & 0.00$\pm$0.00 & 1.06$\pm$0.29 \\
PIA~\cite{kong2023efficient}             & 50.09$\pm$0.56 & 52.29$\pm$0.93 & 0.00$\pm$0.00 & 1.78$\pm$0.35 \\
PIAN~\cite{kong2023efficient}            & 51.69$\pm$0.55 & 50.90$\pm$0.95 & 0.00$\pm$0.01 & 2.18$\pm$0.48 \\
\bottomrule
\end{tabular}
\vspace{-0.1cm}
\end{wraptable}

For DMs, \Cref{tab:dm_mia_metrics} shows AUC and TPR@1\% values for a range of MIA strategies. AudioLDM2 again yields performance near chance, with AUCs around 50–55\%. In contrast, TANGO shows a detectable membership signal, with AUC values reaching nearly 70\%. This difference likely arises from the smaller scale of TANGO’s training set, which makes overfitting more apparent. Results with TPR@FPR=1\%, makes this distinction even clearer: for AudioLDM2 the detection rate is essentially zero across all attacks, while for TANGO it reaches 16–17\% for the best-performing MIAs are feasible only for smaller DMs, but scale poorly to models trained on larger datasets.  

\begin{wraptable}{r}{0.35\linewidth}
\centering
\vspace{-0.4cm}
\scriptsize
\caption{\textbf{results for Dataset inference.} Minimum number of samples required to achieve mean $p\!\le\!0.01$.}
\label{tab:di_samples_wrap}
\begin{tabular}{lc}
\toprule
\textbf{Model} & \textbf{\# Samples} \\
\midrule
AudioGen   & 900 \\
FIGARO     & 300 \\
AudioLDM2  & 300 \\
Tango      & 20 \\
\bottomrule
\end{tabular}
\end{wraptable}

Finally, \Cref{tab:di_samples_wrap} reports the number of samples required for DI to reach statistical significance. Here, the advantage of DI over MIA becomes clear. AudioGen requires around 900 samples to reject the null hypothesis, while FIGARO and AudioLDM2 require only 300 samples. Most strikingly, TANGO requires just 20 samples, showing that DI can detect training set usage with very small collections. Overall, these results highlight that single-sample MIAs are limited when applied to models trained on larger audio datasets, but DI provides strong and scalable evidence, making it a more practical tool for auditing the training sets of generative audio models.
However, for most models, DI requires more samples than an individual artist is likely to possess, especially given the fact that the lenght of AudioCaps sample is 10 secods. These requirements remain attainable for media owners, but highlight the need for further methodological development.

\section{Discussion and Conclusions}

Our study demonstrates that membership inference alone is not a reliable mechanism for verifying whether individual audio samples contributed to the training of large audio generative models. However, when aggregated via dataset inference, the weak per-sample signals accumulate to provide statistically significant evidence of training set inclusion, even with relatively small collections. This highlights DI as a promising tool for creators and auditors seeking to verify copyright misuse.

An important implication of our findings is the responsibility of model providers in enabling meaningful auditing. Currently, many released models do not disclose clear train–test splits or maintain accessible held-out evaluation sets. This lack of transparency makes it challenging to fairly assess privacy leakage, data governance, and copyright compliance. We argue that providers of generative models should adopt the practice of reporting well-defined train/test partitions and reserving clean held-out sets that remain unused during training. Such held-out data would allow independent researchers to systematically study privacy risks, monitor overfitting, and develop robust detection techniques without ambiguity about data provenance.


\small

\section*{Acknowledgments}
This research was supported by the Polish National Science Centre (NCN) within grant no. 2023/51/I/ST6/02854.We gratefully acknowledge Poland's high-performance Infrastructure PLGrid for providing computer facilities and support within computational grant no. PLG/2025/018230.
This work is also partly supported by the MUR FIS2 grant n. FIS-2023-00942 "NEXUS" (cup B53C25001030001), and by Sapienza University of Rome via the Seed of ERC grant "MINT.AI" (cup B83C25001040001).
This work was also supported by the German Research Foundation (DFG) within the framework of the Weave Programme under the project titled "Protecting Creativity: On the Way to Safe Generative Models" with number 545047250. 

\bibliographystyle{unsrt}
\bibliography{main}


\appendix

\clearpage
\section{Details on Membership Inference for Autoregressive Models}

\textbf{Autoregressive Modeling for Audio \emph{and} Symbolic Music.}
Autoregressive generators in the audio domain operate on compact \emph{discrete sequences} rather than raw waveforms. Two prevalent instantiations are: (i) \emph{tokenized time–frequency} representations, where a vector-quantized encoder (e.g., VQ-VAE/VQ-GAN) maps a mel-spectrogram $\mathrm{Mel}(x)$ to a low-resolution latent grid and quantizes each cell to a codebook index; and (ii) \emph{symbolic music} representations (e.g., MIDI or piano-roll), where events such as \textsc{Note-On}, \textsc{Note-Off}, pitch, velocity, instrument, bar/beat markers, and control changes are serialized into a discrete token stream. In both cases, the 2D structure (time–frequency for mel tokens; hierarchical/bar–beat–event for MIDI) is linearized into a 1D sequence $\mathbf{c}=(c_1,\ldots,c_N)$ using a fixed, deterministic ordering (typically time-major; optional bar/measure delimiters for music).

The generative objective models next-token probabilities:
\begin{equation}
p(\mathbf{c}) \;=\; \prod_{n=1}^{N} p\!\left(c_n \mid c_1, \ldots, c_{n-1}\right),
\label{eq:ar_task_tokens}
\end{equation}
optimized via maximum likelihood over the training distribution:
\begin{equation}
L_{\mathrm{AR}} \;=\; \mathbb{E}_{x \sim \mathcal{D}_{\mathrm{train}}}\!\left[-\log p\!\big(\mathbf{c}(x)\big)\right],
\label{eq:ar_loss_tokens}
\end{equation}
where $\mathbf{c}(x)$ denotes either codebook indices derived from $\mathrm{Mel}(x)$ \emph{or} a serialized symbolic/MIDI event stream for $x$.

This token-first formulation shortens effective sequence length and exposes strong discrete structure (repetition, meter, harmony), enabling high-fidelity generation with tractable context windows. At inference, tokens $\hat{\mathbf{c}}$ are sampled autoregressively from Eq.~\ref{eq:ar_task_tokens}. For mel-token systems, a quantized decoder reconstructs a mel-spectrogram $\widehat{\mathrm{Mel}}$ that is rendered to waveform via a vocoder. For symbolic systems (e.g., FIGARO), the sampled event stream is rendered to audio by a MIDI synthesizer or mapped back to a structured score, optionally honoring controllable conditioning tokens (e.g., chords, instrumentation, rhythm descriptors) embedded in the same sequence.

Kowalczuk et al.~\cite{kowalczukprivacy} introduce the first comprehensive MIA suite for \emph{image} autoregressive models by adapting well-established, token-level attacks from the LLM literature (e.g., Loss, Zlib, Hinge, Min-K\%, Min-K\%$^{++}$, SURP, CAMIA) to visual next-token prediction. A key observation in their work is that many IARs are trained with \emph{classifier-free guidance}~\cite{ho2022classifier}, i.e., the forward pass processes each example both \emph{with} conditioning (e.g., a class label or text prompt) and \emph{without} it. Building on CLiD~\cite{zhai2024clid}, they exploit this extra supervision signal by contrasting the conditional and unconditional paths: instead of feeding raw per-token logits into MIAs, they use the guidance-difference statistic
\[
\Delta(x,c) \;=\; p(x \mid c)\;-\;p(x \mid c_{\text{null}}),
\]
where $c$ is the conditioning input and $c_{\text{null}}$ denotes the null (unconditional) condition. This replacement amplifies membership signal relative to LLM-style attacks that lack such conditioning, and it avoids relying solely on per-token probabilities.

Because \emph{audio} autoregressive models also operate on discrete token sequences, either time–frequency codes (e.g., codec/VQ tokens) or symbolic music events (e.g., MIDI as in FIGARO), the same LLM-derived MIAs are directly applicable in the audio domain. When audio ARMs are trained with explicit conditioning (e.g., captions, tags, control tokens) and employ classifier-free guidance, the CLiD-style conditional–unconditional contrast $\Delta(x,c)$ can be computed analogously on audio tokens and used as the primary MIA feature. 

\textbf{Threshold-based attack.}
A simple and widely used approach to infer membership is to compare a scalar diagnostic to a fixed cutoff. Let $\mathcal{M}$ be a per-sample metric such as the negative log-likelihood or loss. A sample $x$ is declared a member whenever the metric falls below a threshold $\gamma$:
\begin{equation}
A_{f_\theta}(x) \;=\; \mathbb{1}\!\left[\mathcal{M}(f_\theta,x) < \gamma\right],
\label{eq:mia_thr}
\end{equation}
where $\gamma$ is selected on a validation split. The rationale is that training items typically attain lower loss than points not seen during training.

\textbf{\mink\ metric.}
To reduce the influence of highly predictable positions, \cite{shi2024detecting} focus the decision rule on the least likely part of the sequence. For an input $x$ and a fraction $K\!\in\!\{10,20,30,40,50\}$, \mink\ computes the average negative log-likelihood over the bottom $K\%$ tokens under the model $f_\theta$. Membership is predicted by thresholding this average:
\[
A_{f_\theta}(x) \;=\; \mathbb{1}\!\left[\operatorname{Min}\text{-}K\%(x) < \gamma\right].
\]
Reporting the best result over a small sweep of $K$ makes the attack less sensitive to the choice of this hyperparameter.

\textbf{\mink++}.
\mink++ refines \mink\ by normalizing token log-probabilities and testing whether low-probability positions behave like local modes of the learned distribution. Given a sequence $x=(x_1,\dots,x_T)$, define
\begin{equation}
\mathcal{S}_{\text{Min-K\%++}}(x) \;=\; \frac{1}{|S|}\sum_{t\in S}\frac{\log p(x_t\mid x_{<t})-\mu_{x<t}}{\sigma_{x<t}},
\end{equation}
where $S$ is the subset containing the bottom $K\%$ tokens, and $\mu_{x<t}$, $\sigma_{x<t}$ are the mean and standard deviation of token log-probabilities over the entire vocabulary at position $t$. A sample is flagged as a member if
\begin{equation}
A_{f_\theta}(x) \;=\; \mathbb{1}\!\left[\mathcal{S}_{\text{Min-K\%++}}(x) \ge \gamma\right].
\end{equation}
As with \mink, performance is reported for the best $K\in\{10,20,30,40,50\}$.

\textbf{Zlib ratio attack.}
This baseline relates model fit to a model-agnostic compressibility proxy~\citep{zlib2004}. Let $\mathcal{P}_{f_\theta}(x)$ denote the perplexity (or exponentiated average negative log-likelihood) and $\operatorname{zlib}(x)$ be the compressed size of $x$ under the zlib codec. The statistic
\[
\frac{\mathcal{P}_{f_\theta}(x)}{\operatorname{zlib}(x)}
\]
tends to be smaller for members, since model perplexity is lower on training data while zlib compression does not benefit from any model-specific memorization. Membership is then inferred by comparing this ratio to a threshold.

\textbf{CAMIA.}
Context-aware MIA~\citep{chang2024context} augments raw loss features with temporal descriptors of the token-wise loss sequence. Several signals are used: a \emph{slope} feature that captures how quickly losses decline across positions; \emph{approximate entropy}, which measures regularity by the prevalence of repeating patterns; \emph{Lempel–Ziv complexity}, which quantifies diversity in the loss trajectory via the count of distinct substrings; a \emph{count-below} statistic, the fraction of tokens with loss below a preset cutoff; and a \emph{repeated-sequence amplification} feature that measures the reduction in loss when the same input is repeated. Non-members typically display higher irregularity and larger gains from repetition, while members show more stable, low-loss segments.

\textbf{Surprising Tokens Attack (SURP).}
SURP targets positions where the model is confident overall but assigns low probability to the true token. For each position $t$, let $H_t$ be the Shannon entropy of the predictive distribution and $p(x_t\mid x_{<t})$ the probability of the ground-truth token. Define the surprising set
\begin{equation}
S \;=\; \left\{\, t \;\middle\vert\; H_t < \epsilon_e,\; p(x_t\mid x_{<t}) < \tau_k \,\right\},
\end{equation}
where $\epsilon_e \in \{2,4,8,16\}$ controls the entropy threshold and $\tau_k$ is the $k$-th percentile probability with $k\in\{10,20,30,40,50\}$. The SURP score averages the probabilities on this set:
\begin{equation}
\mathcal{S}_{\text{SURP}}(x) \;=\; \frac{1}{|S|}\sum_{t\in S} p(x_t\mid x_{<t}).
\end{equation}
Membership is decided by thresholding $\mathcal{S}_{\text{SURP}}(x)$:
\begin{equation}
A_{f_\theta}(x) \;=\; \mathbb{1}\!\left[\mathcal{S}_{\text{SURP}}(x) \ge \gamma\right].
\end{equation}
In practice, the best-performing pair $(k,\epsilon_e)$ from the specified grids is selected to summarize results.

\section{Details on Membership Inference for Diffusion Models}

\textbf{Diffusion Models}~\cite{songdenoising}
are generative models trained by progressively adding noise to the data and then learning to reverse this corruption.
In the forward diffusion process, Gaussian noise $\epsilon \sim \mathcal{N}(0, I)$ is added to a clean sample $x$ to obtain a noised sample
$x_t \gets \sqrt{\alpha_t}\, x + \sqrt{1-\alpha_t}\, \epsilon$, where $t\in[0,T]$ is the diffusion timestep and
$\alpha_t \in [0,1]$ is a monotonically decreasing schedule with $\alpha_0=1$ and $\alpha_T=0$.
The denoiser $f_\theta$ is trained to predict $\epsilon$ across timesteps by minimizing
$\frac{1}{N}\sum_i \mathbb{E}_{t,\epsilon}\,\mathcal{L}(x_i,t,\epsilon; f_\theta)$, where $N$ is the training set size and
\begin{align}
\mathcal{L}(x,t,\epsilon; f_\theta) = \left\lVert \epsilon - f_\theta(x_t, t) \right\rVert_2^2 \, .
\label{eq:pixel_training_loss}
\end{align}

Sampling proceeds by iteratively removing predicted noise $f_\theta(x_t,t)$ from $x_t$ for $t=T,T-1,\ldots,0$, starting from $x_T \sim \mathcal{N}(0,I)$ to obtain a generated sample $x_{t=0}$. For conditional generation, an additional input $y$ (conditioning signal) is provided to $f_\theta$.

Latent diffusion models~\cite{rombach2022high} perform the diffusion process in a learned latent space to improve efficiency. An encoder $\mathcal{E}$ maps $x$ to a latent $z=\mathcal{E}(x)$, and the objective in Eq.~\ref{eq:pixel_training_loss} becomes
\begin{align}
\mathcal{L}(z,t,\epsilon; f_\theta) = \left\lVert \epsilon - f_\theta(z_t, t) \right\rVert_2^2 \, .
\label{eq:training_loss}
\end{align}

\textbf{Denoising Loss.}
Early membership inference attacks for diffusion models~\cite{carlini2023extracting} assess sample membership by directly using the denoising loss as a statistic. The key observation is that the loss at intermediate timesteps provides the strongest separation between training members and non-members. In particular, $t\approx 100$ often yields the most discriminative signal: very small $t$ makes the task too easy (the noised input remains close to the original), whereas very large $t$ collapses the input toward pure noise, making prediction uniformly hard. A sample is classified as a member if its loss at the chosen timestep falls below a threshold selected on a validation split.

\paragraph{Multiple Loss.}
A multi-timestep variant aggregates information from several diffusion steps to improve robustness of the signal. This attack evaluates Eq.~\ref{eq:training_loss} at a fixed grid of timesteps (e.g., $t \in \{0,100,\ldots,900\}$) and combines the resulting losses into a single score, for example by summation or a weighted average. The aggregate loss serves as the decision statistic, again thresholded to yield a membership prediction. Using multiple timesteps reduces variance and can capture complementary difficulty regimes of the denoising task.

\textbf{Proximal Initialization Attack (PIA).}
The PIA family~\cite{kong2023efficient} compares the model’s noise predictions when initialized from different proximity states to the data. A canonical instantiation evaluates the prediction error at a clean (or minimally noised) state, such as $t=0$, and at a moderately noised state, typically around $t=200$ where separability is reported to be strong. The difference (or ratio) between these errors is used as the attack feature. Intuitively, training samples induce more confident and stable predictions across nearby states of the diffusion process, leading to a lower feature value for members than for non-members.

\textbf{PIAN.}
An adaptation of PIA, denoted PIAN~\cite{kong2023efficient}, normalizes the denoiser’s output to enforce approximately Gaussian behavior in the predicted noise, thereby reducing scale effects that may confound raw error magnitudes. The membership statistic is computed analogously to PIA after normalization. As with PIA, members are expected to yield smaller scores because the model’s predictions align more consistently with the true noise on training data.

\paragraph{Gradient Masking.}
The gradient-masking attack~\cite{dubinski2025cdi} targets semantically critical regions of the latent representation that most influence the denoising loss. For a given $z_t$, the gradient $\mathbf{g} = \left\lvert \nabla_{z_t}\, \mathcal{L}(z_t,t,\epsilon; f_\theta) \right\rvert$ is computed, and a binary mask $\mathbf{M}$ is formed by selecting the top-percentile (e.g., top $20\%$) entries of $\mathbf{g}$.
A perturbed latent $\hat{z}_t = \epsilon \cdot \mathbf{M} + z_t \cdot \neg \mathbf{M}$ is then created by replacing the most influential coordinates with random noise and leaving the remainder unchanged. The attack feature is the reconstruction error restricted to the masked region,
$\left\lVert (\epsilon - z_t)\cdot \mathbf{M} - f_\theta(\hat{z}_t, t)\cdot \mathbf{M} \right\rVert_2^2$,
optionally aggregated across multiple timesteps. Because models tend to memorize salient structure in training samples, members exhibit lower masked-region reconstruction error than non-members.

\paragraph{Noise Optimization.}
The central premise of noise-optimisation attack~\cite{dubinski2025cdi} is that stronger (or more effective) perturbations are required to significantly reduce the denoising loss for training members, reflecting higher confidence and tighter fit on seen data. Concretely, starting from $z_t$ at an intermediate timestep (e.g., $t=100$), an unconstrained optimization seeks a perturbation $\delta$ that minimizes the objective
$\min_{\delta}\, \left\lVert \epsilon - f_\theta(z_t + \delta, t) \right\rVert_2^2$,
using 5 L-BFGS steps. Two complementary features arise: the minimized prediction error
$\left\lVert \epsilon - f_\theta(z_t + \delta, t) \right\rVert_2^2$ and the perturbation magnitude $\lVert \delta \rVert_2^2$.
Members typically achieve lower final error yet require larger or more targeted adjustments, producing distinctive signatures relative to non-members.

\section{Details on Dataset Inference}

DI generalizes MIAs from individual samples to sets. Its central research question is: \textit{was the collection of suspect samples $P$ used to train the model, as opposed to being independent test data?} To answer this, DI compares $P$ against a reference set $U$ drawn from the same distribution but known to be excluded from training. In both DMs and IARs the procedure consists of three steps: (i) extract a suite of per-sample MIA features, (ii) map these features into scalar membership scores, and (iii) perform a statistical test comparing the distributions of scores for $P$ and $U$. The null hypothesis $H_0:\text{mean}(s(P)) \leq \text{mean}(s(U))$ is tested with Welch’s $t$-test at $\alpha=0.01$.

\textbf{Diffusion models.}  
For DMs, the CDI methodology~\cite{dubinski2025cdi} employs a broad feature set, which we describe in Appendix~B. Rather than aggregating these features directly, CDI fits a logistic regression scorer on disjoint control splits of $P$ and $U$, yielding a calibrated mapping from feature vectors to scalar scores. The test is then applied to scores on held-out subsets. To reduce variance, CDI repeats this process across multiple random partitions and averages the resulting $p$-values.

\textbf{Image autoregressive models.}  
For IARs, the approach of~\cite{kowalczukprivacy} follows the same overall structure but makes use of a different feature suite, tailored to token-level modeling. These features, described in Appendix~A, capture variations in token probabilities and loss trajectories that arise in autoregressive generation. Each feature is normalized, and the per-sample scalar score is obtained by summing across all features. This lighter-weight procedure that nevertheless suffices in practice for autoregressive token-based models. The resulting scores for $P$ and $U$ are then compared using the same statistical test as above.


\end{document}